\documentclass[12pt]{article}
\usepackage{amsmath,amssymb,amsthm}
\usepackage{booktabs}
\usepackage{graphicx}
\usepackage{hyperref}

\title{Risk-Entropic Flow Matching }
\author{
  Vahid R. Ramezani\thanks{Email: \texttt{vahid@engramsciences.com}} \and
  Benjamin Englard\thanks{Email: \texttt{ben@engramsciences.com}}
}
\date{November 2025}

\theoremstyle{plain}
\newtheorem{theorem}{Theorem}

\newtheorem{lemma}{Lemma}
\theoremstyle{definition}

\theoremstyle{remark}

\newtheorem{corollary}{Corollary}

\begin{document}
\sloppy
\maketitle

\begin{abstract}
Tilted (entropic) risk, obtained by applying a log-exponential transform to a
base loss, is a well established tool in statistics and machine learning for
emphasizing rare or high loss events while retaining a tractable optimization
problem. In this work, our aim is to interpret its structure for Flow Matching(FM). FM learns a velocity field that transports samples from a simple source distribution to data
by integrating an ODE. In rectified FM, training pairs are obtained by linearly
interpolating between a source sample and a data sample, and a neural velocity
field is trained to predict the straight line displacement using a mean squared
error loss. This squared loss collapses all velocity targets that reach the same
space-time point into a single conditional mean, thereby ignoring higher order
conditional information (variance, skewness, multi-modality) that encodes
fine geometric structure about the data manifold and minority branches.

We apply the standard risk-sensitive (log-exponential) transform to the conditional FM loss and show that the resulting tilted risk loss is a natural upper-bound on a
meaningful conditional entropic FM objective defined at each space-time
point. Furthermore, we show that a small order expansion of the gradient of this conditional entropic objective yields two interpretable first order corrections: covariance
preconditioning of the FM residual, and a skew tail term that favors asymmetric
or rare branches. On synthetic data
designed to probe ambiguity and tails, the resulting risk-sensitive loss improves statistical metrics and recovers geometric structure more faithfully
than standard rectified FM.

\end{abstract}

\section{Introduction}

Flow matching (FM) has recently emerged as a powerful alternative to
score based diffusion models for building continuous time generative models.
Most formulations start from a prescribed probability path $(p_t)_{t\in[0,1]}$
and construct a vector field $u_t(x)$ such that the associated probability flow
ODE transports a simple source distribution $p_0$ to a complex target
$p_1$~\cite{lipman2022flowmatching,lipman2023,guo2025vrfm}.
A neural network $u_t^\theta(x)$ is trained by minimizing a supervised squared
loss between the model field and a microscopic target velocity defined
along conditional paths. 

 The training objective is an $L^2$ regression problem in which a model
vector field $v_t(x)$ is fit to a random microscopic velocity
$u_t(X_t \mid X_1)$ under the path measure. Formally, classical FM solves,
at each $(t,x)$, the $L^2$ projection problem
\[
u_t(x)
\;=\;
\mathbb{E}\big[u_t(X_t \mid X_1)\,\big|\,X_t = x\big],
\]
so that the target velocity field is the conditional mean of a richer ensemble
of conditional paths. The associated ODE or probability flow construction then
serves as a mechanism for aggregating these local regression problems into a
global generative model with the correct marginals in the idealized limit of a
perfectly fitted vector field.

This is not to say that FM, as a generative framework, is merely a regression estimator. 
Flow matching has recently been shown to achieve almost minimax optimal convergence rates 
under strong smoothness and regularity assumptions on the data distribution and the 
interpolation path \cite{albergo2023stochastic,benton2024error, fukumizu2024flow}. However, these guarantees  typically rely on globally smooth, light-tailed densities, globally Lipschitz velocity fields, 
and strong regularity  that are hard to verify in practice and can lead to very loose constants in high dimension.  Real-world data manifolds are highly complex and far from globally smooth. Empirical studies of natural-image statistics consistently report sharply peaked, heavy-tailed filter-response  distributions \cite{ruderman1994woods,huang1999stats}. Furthermore, the bounds, naturally, depend on the size of datasets.

 In many areas of statistics, control, and finance, it is well understood that one can move beyond purely mean based criteria by replacing the plain expectation of a loss with a
\emph{ risk-sensitive} or \emph{tilted} risk \cite{lee2020learningsensitive} \cite{li2021term} \cite{sypherd2022tunable}, for example via the log-exponential (entropic) transform as in \cite{howard1972,ramezani2002hmm,detlefsen2005conditional,follmer_schied_book, leqi2022generalrisk} and   recent work on tilted empirical risk in machine learning~\cite{li2021term,aminian2025tilted}.

This endows the same underlying loss with a different aggregation functional
that emphasizes variability, tails, or rare events, and has been studied both
for its decision theoretic properties and for its robustness in learning. What is its interpretation for FM?  Answering this question is the main contribution of this paper. 

In this work, we apply this entropic risk principle to the FM loss, and show
that the resulting structure is particularly tractable. At the local level, we
introduce a \emph{conditional entropic flow matching objective} obtained by
applying a log-exponential transform to the squared FM residual
$\|u_t^\theta(x)-U_t(x,z)\|^2$. The gradient of this objective retains a clean FM form: it can be written as the usual FM update with the
target velocity replaced by a Gibbs tilted conditional mean $m_t^\lambda(x;\theta)$.
A small $\lambda$ expansion of this tilted mean reveals two interpretable
first order corrections to standard FM: a covariance preconditioning term that
rescales the FM residual by the conditional covariance of the microscopic
velocity, and a skew/tail term that biases updates towards asymmetric or rare
branches of the conditional path ensemble. Furthermore, optimizing the conditional entropic objective directly is computationally
inconvenient, but it admits the marginal \emph{tilted risk} as an upper-bound.

We begin with some background on rectified Flow Matching following the development given in \cite{guo2025vrfm}. Assume $p_0(\xi_0)=\mathcal{N}(\xi_0; x_0, I)$ and $p_1(\xi_1)=\mathcal{N}(\xi_1; x_1, I)$ with fixed means $x_0,x_1 \in \mathbb{R}^d$, and consider a constant velocity field $v_\theta(\xi_t,t)\equiv \theta$. The likelihood of a data point $x_1$ can be assessed via the instantaneous change of variables formula~\cite{lipman2023}
\begin{equation}
\log p_1(x_1) = \log p_0(x_0)
  + \int_0^1 \operatorname{div} v_{\theta}(x_t, t)\, dt.
\label{eq:instantaneous-cov}
\end{equation}
Since $\operatorname{div} v_\theta \equiv 0$, the instantaneous change of variables formula~\eqref{eq:instantaneous-cov}  reduces to
\begin{equation}
\log p_1(\xi_1) - \log p_0(\xi_0) = 0,
\end{equation}
on the other hand, by definition:
\begin{equation}
\xi_1 = \xi_0 + \int_0^1 v_\theta(\xi_t,t)\,dt = \xi_0 + \theta .
\label{eq:gaussian-cov-constant}
\end{equation}
Using the Gaussian forms of $p_0$ and $p_1$, the identity \eqref{eq:gaussian-cov-constant} is equivalent to
\begin{equation}
(\xi_0 - x_0)^\top(\xi_0 - x_0)
-
(\xi_1 - x_1)^\top(\xi_1 - x_1)
\equiv 0
\qquad \forall\,\xi_0,\xi_1 .
\end{equation}
Substituting $\xi_1 = \xi_0 + \theta$ yields, for all $\xi_0 \in \mathbb{R}^d$,
\begin{align}
0
&= (\xi_0 - x_0)^\top(\xi_0 - x_0)
   - (\xi_0 + \theta - x_1)^\top(\xi_0 + \theta - x_1) \\
&= (x_1 - x_0 - \theta)^\top \bigl( 2\xi_0 - x_0 - x_1 + \theta \bigr).
\end{align}
Since this must hold for every $\xi_0$, we must have $x_1 - x_0 - \theta = 0$, i.e.
\begin{equation}
\theta = x_1 - x_0.
\end{equation}
Consequently, the flow induced by the ODE $\dot x_t = v_\theta(x_t,t) \equiv \theta$ is
\begin{equation}
x_t = x_0 + t\theta = (1-t)x_0 + t x_1,
\end{equation}
which is precisely the rectified (linear) interpolation between $x_0$ and $x_1$.

In this uni-modal setting, it is natural to model the  ground truth velocity as a Gaussian observation
\begin{equation}
x_1 - x_0 \;\sim\; \mathcal{N}\bigl(v_\theta(x_t,t), I\bigr),
\end{equation}
so that the negative log-likelihood is, up to an additive constant,
\begin{equation}
\mathcal{L}(\theta)
= \mathbb{E}\Bigl[\bigl\|v_\theta(x_t,t) - (x_1 - x_0)\bigr\|_2^2\Bigr].
\end{equation}
Thus, the usual squared loss used in rectified flow matching coincides with maximum likelihood for this Gaussian observation model and is therefore principled in this unimodal case.

The simple setup provides insight into the objective of classic rectified flow matching and offers an alternative way to interpret the flow matching procedure. Instead of two single Gaussian probability density, we now assume that the true source and target distributions are empirical mixtures of Gaussians centered at the data points. For example, for a finite dataset $\mathcal{S}\subset\mathbb{R}^d$ we may write
\begin{equation}
p_0(\xi_0)
=\frac{1}{|\mathcal{S}_0|}\sum_{x_0\in\mathcal{S}_0} \mathcal{N}(\xi_0;x_0,I),
\qquad
p_1(\xi_1)
=\frac{1}{|\mathcal{S}_1|}\sum_{x_1\in\mathcal{S}_1} \mathcal{N}(\xi_1;x_1,I),
\end{equation}
with $\mathcal{S}_0,\mathcal{S}_1$ the source and target samples respectively.

Let's assume that the velocity vector field $v_\theta(x_t,t)$ at a space-time location $(x_t,t)$ is characterized by a unimodal Gaussian observation model
\begin{equation}
p\bigl(v \mid x_t,t\bigr)
=
\mathcal{N}\bigl(v; v_\theta(x_t,t), I\bigr),
\end{equation}
with parametric mean $v_\theta(x_t,t)$ and unit covariance. In the rectified flow setting, a pair $(x_0,x_1)$ is connected by the linear trajectory
\begin{equation}
x_t = (1-t)x_0 + t x_1,
\end{equation}
so that the associated empirical velocity is simply $x_1 - x_0$, which is independent of $t$ for a given pair. Maximizing the log-likelihood of these empirical velocity observations at $(x_t,t)$ leads to the objective
\begin{equation}
\mathbb{E}_{t,x_0,x_1}\bigl[\log p(x_1 - x_0 \mid x_t,t)\bigr]
\;\propto\;
-\mathbb{E}_{t,x_0,x_1}\bigl\| v_\theta(x_t,t) - (x_1 - x_0)\bigr\|_2^2,
\end{equation}
where the expectation is over $t\sim\mathcal{U}[0,1]$ and data pairs $(x_0,x_1)$. Thus, up to an additive constant and a positive scale factor, the negative log-likelihood coincides with the classic rectified flow matching loss:
\begin{equation}
\mathcal{L}_\mathrm{RF}(\theta)
\;=\;
\mathbb{E}_{t,x_0,x_1}\bigl\| v_\theta(x_t,t) - (x_1 - x_0)\bigr\|_2^2.
\end{equation}
In other words, the standard squared loss used in rectified flow matching can be seen as a principled maximum likelihood estimator for the (Gaussian) observation model on velocities. This interpretation also makes clear that the MSE objective is tailored to the case where the conditional law of the true velocity $x_1 - x_0$ at $(x_t,t)$ is effectively unimodal and Gaussian. In regimes where the conditional distribution is multimodal, heteroscedastic, or skewed, the MSE fit tends to average over distinct plausible directions, yielding a compromise vector field that can under-represent rare but valid branches.

The rest of the paper is organized as follows. First, we briefly review rectified flow matching. Next, we introduce an application of the entropic risk measure ~\cite{ramezani2002hmm,howard1972} to the FM loss. We demonstrate that a first order expansion in the risk coefficient (small $\lambda$) corresponds to a modification of the gradient of the FM parameterized vector field. This conditional formulation, while theoretically grounded, does not lend itself directly to efficient sampling for training. Fortunately, a closely related unconditional risk-sensitive loss is easier to sample from and we adopt this practical surrogate in our experiments.

\section{Background: Flow Matching and Rectified Paths}

Let $(\mathcal{X},\langle\cdot,\cdot\rangle)$ be Euclidean with norm $\|\cdot\|$. A (time dependent) velocity field
$u_\theta^t:\mathcal{X}\to\mathcal{X}$ induces a path of probability measures $(p_t)_{t\in[0,1]}$ via the continuity equation
\begin{equation}
\partial_t p_t(x) + \nabla_x\!\cdot\!\big(p_t(x)\,u_\theta^t(x)\big)=0,
\label{eq:continuity}
\end{equation}
with prescribed source $p_0$ and terminal distribution $p_1$ matching the data.

Conditional Flow Matching (CFM)  specifies a training distribution over space-time points and local velocity targets.  We will use $z$ in the rest of this paper for data points used for the construction of conditional paths.   In the flow-matching literature, \(U_t(x,z)\) is often written as \(U_t(x\mid z)\) and is referred to as a conditional vector field. However, the condition is just a reminder that   \(U_t(x\mid z)\)  is the vector field that generates \(p_t(\cdot |z)\) and not an actual probabilistic condition.   The CFM objective is the mean-squared regression, which we denote with MSE throughout:
\begin{equation}
\mathcal{L}_{\mathrm{MSE}}(\theta)
=
\mathbb{E}\Big[\;\big\|u_\theta^t(x)-U_t(x,z)\big\|^2\;\Big],
\label{eq:cfm}
\end{equation}
which trains the parametric field $u_\theta^t$ to match the conditional velocity field $U_t(x,z)$ along the path $p_t$. Rectified Flow Matching  is obtained by a particular choice of training pair construction defining a constant velocity $U_t(x,z)= x_1- x_0$: 

\begin{equation}
\mathcal{L}_{\mathrm{FM}}(\theta)
=
\mathbb{E}_{t,x_0,x_1}\!
\Big[\;\big\|u_\theta^t(x_t)-\big(x_1-x_0\big)\big\|^2\;\Big],
\label{eq:rfm-loss}
\end{equation}

In Conditional Flow Matching, the conditional mean velocity \(\mu_t(x)=\mathbb{E}[\,U_t \mid X_t{=}x]\)  is  the transport field that renders the path \(p_t\) a solution of the continuity equation. Our risk tilted objective changes only the \emph{local} averaging rule, as shown below,  the same gradient form appears with \(\mu_t\) replaced by a tilted mean.

\section{Risk-Sensitive Flow Matching}

In this section, we introduce Risk-Sensitive Flow Matching. We show that for small values of risk-sensitive/temperature parameter $\lambda$, the risk loss perturbation amounts to a local higher order perturbation of the velocity field.

Define the centered residual
\begin{equation}
\varepsilon_t\;:=\;U_t-\mu_t(X_t),
\qquad
\mathbb{E}[\varepsilon_t\mid X_t]=0.
\label{eq:residual}
\end{equation}
where 
\begin{equation}
\mu_t(x)\;:=\;\mathbb{E}[\,U_t\mid X_t=x],
\label{eq:cond-mean}
\end{equation}

\paragraph{Gradient of the mean squared objective.}
Consider the squared loss
\begin{equation}
\mathcal{L}_{\mathrm{MSE}}(\theta)
=\mathbb{E}\bigl[\ \|u_\theta^t(X_t)-U_t\|^2\ \bigr],
\label{eq:mse}
\end{equation}
where $u_\theta^t:\mathcal{X}\to\mathbb{R}^d$ is differentiable in $\theta$ and integrable so that differentiation under the
integral is valid. Writing $u_\theta:=u_\theta^t$ for brevity, the chain rule gives
\[
\nabla_\theta \|u_\theta(X_t)-U_t\|^2
=2\,(u_\theta(X_t)-U_t)^\top \nabla_\theta u_\theta(X_t).
\]
Taking expectations and applying the tower property,
\begin{align*}
\nabla_\theta \mathcal{L}_{\mathrm{MSE}}(\theta)
&=2\,\mathbb{E}\big[(u_\theta(X_t)-U_t)^\top \nabla_\theta u_\theta(X_t)\big]\\
&=2\,\mathbb{E}\Big[\,\mathbb{E}\big[(u_\theta(X_t)-U_t)^\top \nabla_\theta u_\theta(X_t)\ \big|\ X_t,\ t\big]\Big].
\end{align*}
Conditioning on $(X_t=x)$ and using $U_t=\mu_t(x)+\varepsilon_t$ with $\mathbb{E}[\varepsilon_t\mid X_t=x]=0$,
\[
\mathbb{E}\big[(u_\theta(x)-U_t)^\top \nabla_\theta u_\theta(x)\ \big|\ X_t=x\big]
=(u_\theta(x)-\mu_t(x))^\top \nabla_\theta u_\theta(x).
\]
Therefore,
\begin{equation}
\nabla_\theta \mathcal{L}_{\mathrm{MSE}}(\theta)
=2\,\mathbb{E}\big[(u_\theta^t(X_t)-\mu_t(X_t))^\top \nabla_\theta u_\theta^t(X_t)\big].
\label{eq:mse-grad}
\end{equation}

\paragraph{Stationarity and mean collapse.}
In the unconstrained (nonparametric) function class, the global minimizer of $\mathcal{L}_{\mathrm{MSE}}$ is 
\[
u^\star(x,t)=\mu_t(x)\quad\text{a.s.}
\]

\subsection{Risk-Sensitive Objective}

The log-exponential (entropic) transformation of a nonnegative loss $C(\theta)$ is defined as:
\begin{equation}
\mathcal{L}_\lambda(\theta)
\;:=\;
\frac{1}{\lambda}\,\log\,\mathbb{E}\!\left[\exp\!\big(\lambda\,C(\theta)\big)\right],
\qquad \lambda>0.
\label{eq:entropic-def}
\end{equation}
It smoothly interpolates between the standard mean loss  and a soft maximum. This follows from the convexity and smoothness of the log-sum-exp function and its role as the cumulant generating transform.

\eqref{eq:entropic-def} admits an information theoretic representation~\cite[pp.~174]{follmer_schied_book}. Let $P$ be the data distribution and $Q\!\ll\!P$ any reweighting. Then
\begin{equation}
\frac{1}{\lambda}\log\,\mathbb{E}_{P}\!\big[e^{\lambda C}\big]
\;=\;
\sup_{Q\ll P}\left\{\,\mathbb{E}_{Q}[C]\;-\;\frac{1}{\lambda}\,D_{\mathrm{KL}}(Q\|P)\right\}.
\label{eq:dv}
\end{equation}
Thus minimizing $\mathcal{L}_\lambda$ is equivalent to minimizing the worst case expected loss over all KL bounded reweightings of $P$. As $\lambda$ increases, the optimizer $Q^\star$ tilts toward high loss samples; as $\lambda\downarrow 0$, the KL penalty dominates and $Q^\star\!\to\!P$, recovering the usual mean loss. In risk theory, $\mathcal{L}_\lambda$ is the entropic risk measure
(exponential utility certainty equivalent), adding variance and tail sensitive emphasis as $\lambda$ grows.

In our setting:
\[
\mathcal{L}_\lambda(\theta)
=
\frac{1}{\lambda}
\log \mathbb{E}
  \big[
    \exp\!\big(\lambda \,\|u_\theta^t(x)-U_t(x,z)\|^2\big)
  \big]
.
\]
We will show that this loss is an upper bound for a conditional entropic  objective in the next section.

\subsection{Conditional (Entropic/Gibbs) Objective}

Observe that we can write the Conditional Flow Matching loss by reversing the order of conditioning:
\begin{equation}
\begin{aligned}
L_{\mathrm{MSE}}(\theta)
&= \mathbb{E}_{t,z}\Big[
      \mathbb{E}_{x \mid t,z}
      \,\big\|u_\theta^t(x)-U_t(x,z)\big\|^2
    \Big] \\
&= \mathbb{E}_{t,x,z}\big[
      \|u_\theta^t(x)-U_t(x,z)\|^2
    \big] \\
&= \mathbb{E}_{x,t}\Big[
      \mathbb{E}_{z \mid x,t}
      \,\big\|u_\theta^t(x)-U_t(x,z)\big\|^2
    \Big].
\end{aligned}
\label{eq:cfm-loss}
\end{equation}

Starting from the last (conditional) form, we replace the inner conditional mean over $z \mid x$ by a log-exp (entropic) aggregation and for $\lambda>0$, define the conditional risk tilted loss \cite{detlefsen2005conditional}     \cite{howard1972}
\begin{equation}
L_\lambda(\theta)
:=\mathbb{E}_{x,t}\left[\frac{1}{\lambda}\log \mathbb{E}_{z\mid x,t}
\exp\!\Big(\lambda\,\|u_\theta^t(x)-U_t(x,z)\|^2\Big)\right],
\label{eq:logexp}
\end{equation}
The inner log-exponential is the entropic (Gibbs) risk of the squared error at $(x,t)$; it reduces to the mean as $\lambda\to 0$. This form gives us better insight as to how the higher order moments of the microscopic velocity field manifest themselves. Next, we will give the main results. The detailed proofs are given in Appendix B. 

\paragraph{Tilted weights and gradient.}
Define the conditional Gibbs weights
\[
\beta_\lambda(z\mid x,t;\theta):=\frac{\exp(\lambda\,\|u_\theta^t(x)-U_t(x,z)\|^2)}{\mathbb{E}_{z\mid x,t}\exp(\lambda\,\|u_\theta^t(x)-U_t(x,z)\|^2)},\quad \mathbb{E}_{z\mid x,t}[\beta_\lambda]=1,
\]
and the resulting \emph{tilted conditional mean}
\begin{equation}
m_\lambda^t(x;\theta):=\mathbb{E}_{z\mid x,t}\big[\beta_\lambda(z\mid x,t;\theta)\,U_t(x,z)\big].
\label{eq:tilted-mean}
\end{equation}
Differentiating under the integral sign yields
\begin{equation}
\nabla_\theta L_\lambda(\theta)
=
2\,\mathbb{E}_{x,t}\Big[(u_\theta^t(x)-m_\lambda^t(x;\theta))^\top \nabla_\theta u_\theta^t(x)\Big].
\label{eq:logexp-grad2}
\end{equation}
Comparing \eqref{eq:mse-grad} and \eqref{eq:logexp-grad2}, the only change from MSE is the replacement
\[
\mu_t(x)\quad\longmapsto\quad m_\lambda^t(x;\theta)
\]
in the local averaging rule at each $(x,t)$.

\paragraph{Moment expansion of the tilted mean.}
Write
\[
U_t(x,z)=\mu_t(x)+\varepsilon,\qquad \mathbb{E}[\varepsilon\mid x,t]=0,
\]
and define conditional moments
\[
\Sigma_t(x)=\mathbb{E}[\varepsilon\,\varepsilon^\top\mid x,t],\qquad S_t(x)=\mathbb{E}[\varepsilon\,\|\varepsilon\|^2\mid x,t].
\]
Let the residual error be $m_t(x)=u_\theta^t(x)-\mu_t(x)$,  Theorem \ref{thm:tilted-FM-expansion} proven in Appendix B shows
\begin{equation}
m_\lambda^t(x;\theta)=\mu_t(x)+\lambda\,S_t(x)-2\lambda\,\Sigma_t(x)\,m_t(x)+O(\lambda^2).
\label{eq:tilted-expansion}
\end{equation}
from \eqref{eq:logexp-grad2} and \eqref{eq:mse-grad},
\begin{equation}
\nabla_\theta L_\lambda-\nabla_\theta \mathcal{L}_{\mathrm{MSE}}
=
2\,\mathbb{E}_{x,t}\!\left[\big(-\lambda\,S_t(x)+2\lambda\,\Sigma_t(x)\,m_t(x)\big)^\top \nabla_\theta u_\theta^t(x)\right]+O(\lambda^2).
\label{eq:delta-final}
\end{equation}

The covariance term $2\lambda\,\Sigma_t(x)\,m_t(x)$ acts as a \emph{covariance preconditioner}, scaling updates along
directions of large conditional variance and allocating more learning capacity to ambiguous or hard axes. The skew term $\lambda\,S_t(x)$ acts as a \emph{skew bias}, tilting toward minority or tail branches when the conditional distribution is asymmetric. We can think of this perturbation of the gradient as a form of self-guidance, where, in place of conditioning to modify  the underlying probability, we are modifying the probability measure to account for higher order moments of the distribution. 

\paragraph{Risk-Sensitive Objective as an upper bound} 
We now show that risk-sensitive objective upper bounds conditional entropic objective. Using tower property of expectations we can write
\[
\mathcal{L}_\lambda(\theta)
=
\frac{1}{\lambda}
\log \mathbb{E}
  \big[
    \exp\!\big(\lambda \,\|u_\theta^t(x)-U_t(x,z)\|^2\big)
  \big]
.
\]

\[
\mathcal{L}_\lambda(\theta)
=
\frac{1}{\lambda}
\log \mathbb{E}_{x,t}\Big[
  \mathbb{E}_{z \mid x,t}\big[
    \exp\!\big(\lambda \,\|u_\theta^t(x)-U_t(x,z)\|^2\big)
  \big]
\Big].
\]
Compared to \eqref{eq:logexp}, local higher order moments with respect to $z$ are averaged over $(x,t)$ before they are flattened by the logarithm. However, on average, the objective still relies on the minimization of local higher order structure represented by the inner expectation; the only difference is that the logarithm is applied after this averaging rather than before it. We now make this precise.

Define
\[
\phi_\lambda(x,t;\theta)
\;:=\;
\frac{1}{\lambda}\log \mathbb{E}_{z\mid x,t}
\exp\!\big(\lambda\,\|u_\theta^t(x)-U_t(x,z)\|^2\big),
\]
so that the conditional objective is
\[
L_\lambda(\theta)
= \mathbb{E}_{x,t}\big[\phi_\lambda(x,t;\theta)\big],
\]
Since $\log$ is concave, Jensen's inequality for concave functions \cite{dekking2005modern} gives
\[
L_\lambda(\theta)
=
\mathbb{E}_{x,t}[\phi_\lambda(x,t;\theta)]
\;\le\;
\mathcal{L}_\lambda(\theta),
\]
for all $\lambda>0$. Thus, $\mathcal{L}_\lambda$ is a pointwise (in parameters)  upper-bound on the conditional entropic objective $L_\lambda$, and in what follows we simply use $\mathcal{L}_\lambda$ as a tractable surrogate, in the same spirit as such bound surrogates are commonly optimized in machine learning.

\section{Related Work}

To our knowledge, this is the first work to characterize  entropic risk in FM setting. We adopt a tilted (log-exponential) risk on the standard FM loss, building on a long line of work on entropic and risk-sensitive objectives in control and mathematical finance~\cite{howard1972,ramezani2002hmm,detlefsen2005conditional,follmer_schied_book}.  In machine learning, risk-sensitive learning has been studied in a general way by Lee et al.~\cite{lee2020learningsensitive}, who analyze learning bounds for objectives defined via optimized certainty equivalents, including the entropic risk as a special case; Li et al.~\cite{li2021term} introduce \emph{Tilted Empirical Risk Minimization}, applying exponential tilting directly to the empirical loss distribution to emphasize high or low loss samples; Curi et al.~\cite{curi2020adaptivesampling} develop an adaptive sampling algorithm that improves performance on a given fraction of most difficult examples.  Wu et al.~\cite{wu2023understandingdro} reinterpret contrastive learning objectives such as InfoNCE as distributionally robust optimization, with the temperature parameter acting as a Lagrange multiplier that controls the size of an adversarial distribution set.

A line of recent work on rectified flows analyzes the ensemble of transport paths
induced by random source-target couplings and highlights that naive $\ell_2$
averaging of velocity targets can be harmful. Variational Rectified Flow
Matching~\cite{guo2025vrfm} makes this critique explicit and introduces latent
variables to represent multi-modal velocity fields at each $(x_t,t)$, allowing
distinct branches to cross while remaining consistent with the same marginals.
 \cite{zhang2025towards}  introduces a hierarchical system of ordinary differential equations enabling a more accurate modeling of the underlying velocity distribution. Our work is complementary. Combining such mode aware parameterizations with our risk-sensitive objective that  shifts the statistics is an interesting direction for future work.

\section{Experiments on Synthetic Data}

Our early numerical analysis on synthetic data has been encouraging, we share one of them below, more details can be found in the appendix. For ground truth, we replicated the 2D experiment given in \cite{guo2025vrfm}

\begin{figure}[t]
    \centering
    \includegraphics[width=0.6\textwidth]{gt_trajs.png}
    \caption{Ground truth straight line transport paths in the synthetic 2D ring
    experiment. Blue dots mark source points $x_0$ and orange dots their paired
    targets $x_1$ on the ring; each colored segment shows the straight line
    trajectory $x_t = (1-t)x_0 + t x_1$. The outer dashed circle indicates the target
    ring.}
    \label{fig:gt-ring}
\end{figure}

\paragraph{Setup.}
We consider a rectified flow matching task on a synthetic two ring Gaussian mixture:
sources $x_0$ are sampled on an inner ring and targets $x_1$ on an outer ring,
each with $K{=}6$ equally spaced angular lobes and small radial noise. Training
uses pairs $(x_0,x_1)$, a uniform $t\!\sim\!\mathrm{Unif}[0,1]$, rectified
states $x_t=(1-t)x_0+t x_1$, and the supervised velocity $u_t=x_1-x_0$. Thus,
the ground truth transport consists of straight line chords between paired
points on the two rings, as visualized in Figure~\ref{fig:gt-ring}. We compare
standard FM (MSE) to the \emph{marginal} entropic FM objective with a scheduled
risk coefficient $\lambda(t_{\text{step}})$ that linearly ramps from $0$ to
$\lambda_{\max}$ over $500$ steps and is then held fixed. An MLP with
sinusoidal embeddings predicts $v_\theta(x,t)$; all runs use identical seeds,
optimizer (AdamW), batch size, and step budget.

\paragraph{Metrics.}
We report four diagnostics:
(i) $\mathrm{RMSE}_\sigma$ between the model-recovered lobe spread and the
ground truth spread (lower is better);
(ii) \emph{gap violation rate}---the fraction of generated angles whose
nearest-lobe residual exceeds a calibrated threshold (lower is better);
(iii) the mean 1D Wasserstein--1 distance between the empirical distributions
of absolute angular residuals for the ground truth and the model,
$\mathrm{W}_1(|r|)$ (lower is better); and
(iv) the model's estimated spread $\sigma_{\text{model}}$ (for context).

\begin{figure}[t]
    \centering
    \includegraphics[width=0.55\textwidth]{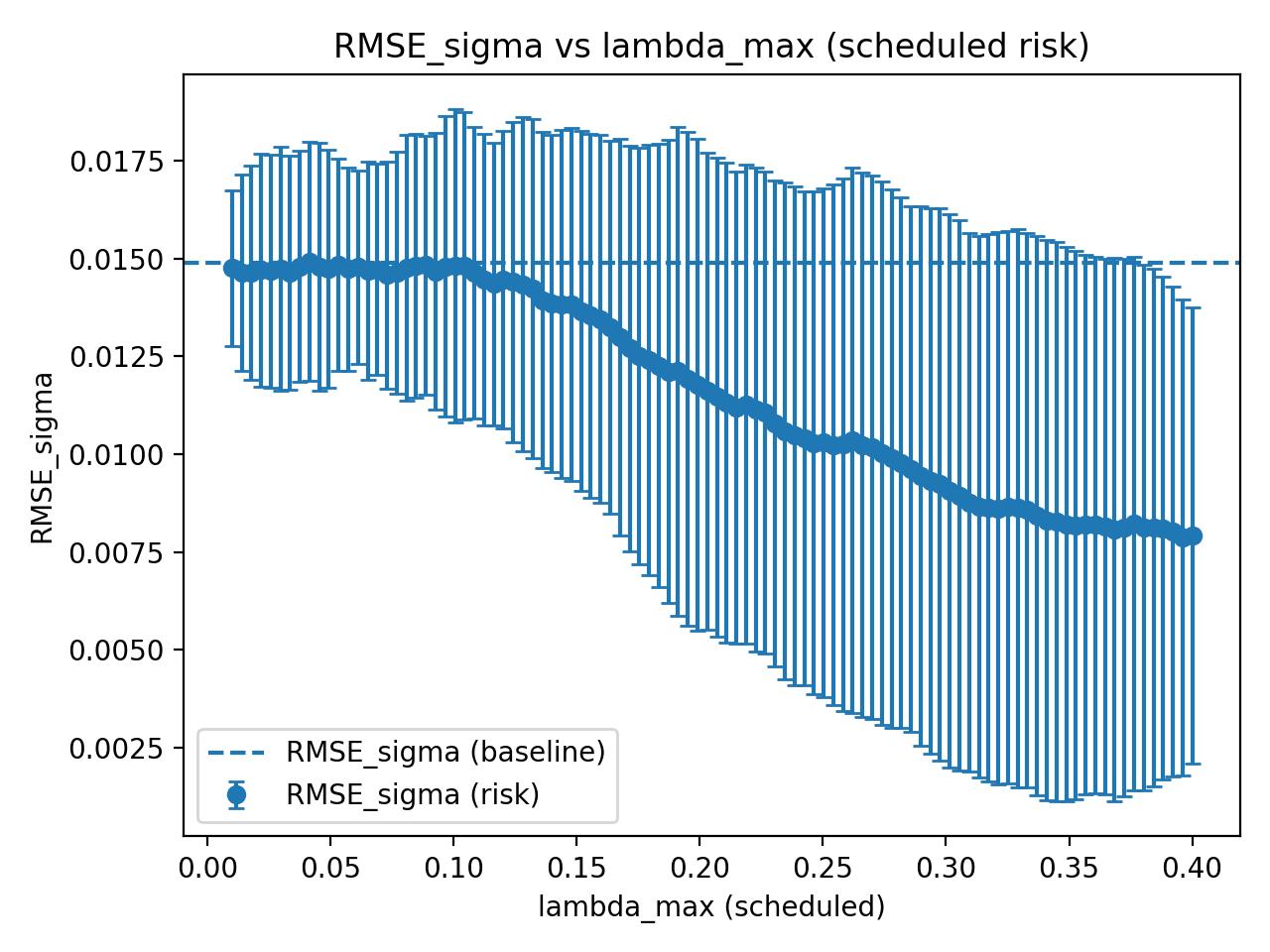}
    \caption{$\mathrm{RMSE}_\sigma$ vs.\ $\lambda_{\max}$ (FM baseline (dashed line) vs.\ scheduled FM-RISK. Lower is better).}
    \label{fig:rmse-sched-raw}
\end{figure}

\begin{figure}[t]
    \centering
    \includegraphics[width=0.55\textwidth]{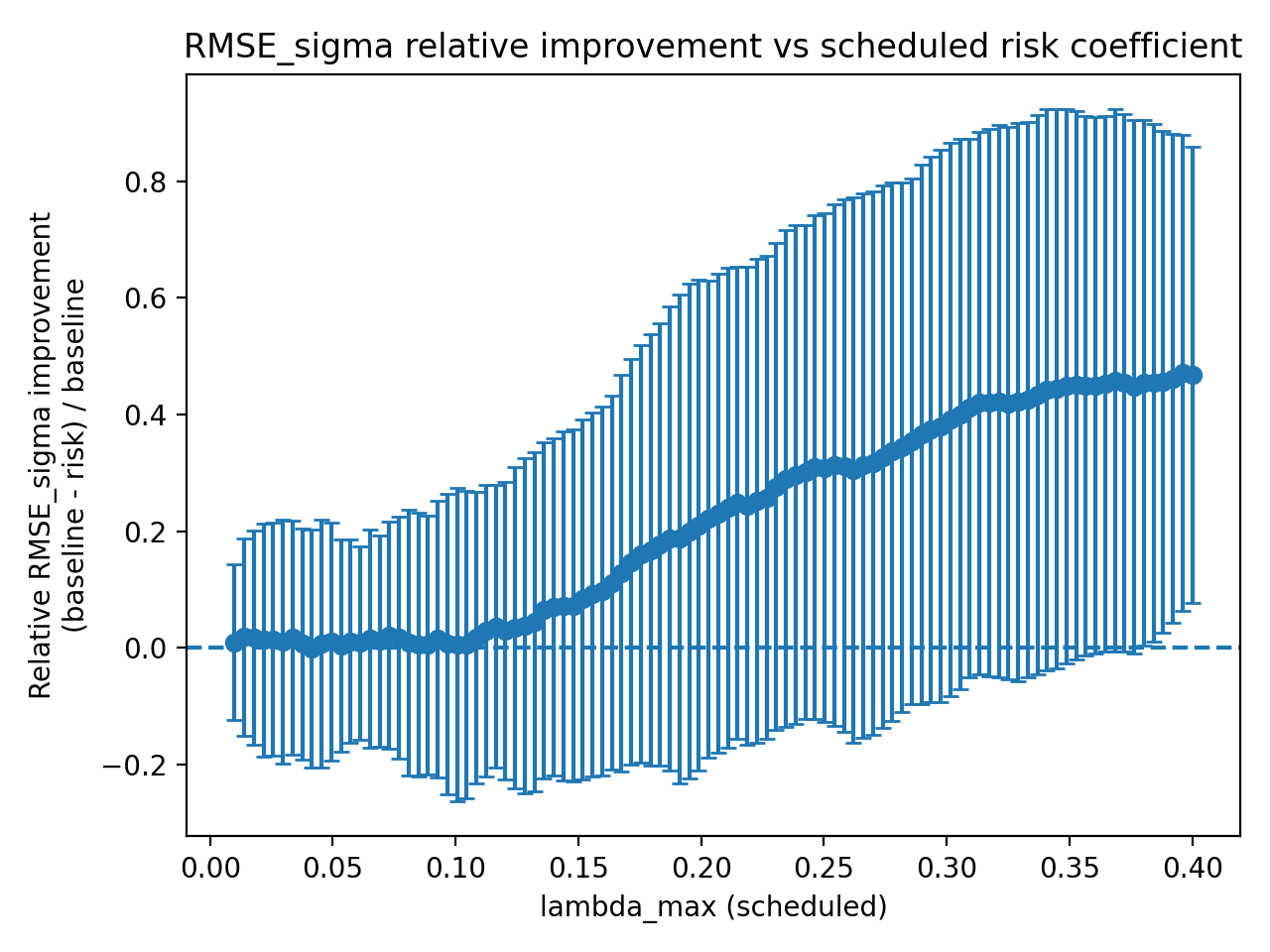}
    \caption{Relative $\mathrm{RMSE}_\sigma$ improvement
    $(\mathrm{RMSE}_\sigma^{\text{base}} -
      \mathrm{RMSE}_\sigma^{\text{risk}}) /
     \mathrm{RMSE}_\sigma^{\text{base}}$ vs.\ $\lambda_{\max}$(FM baseline (dashed line) vs.\ scheduled FM-RISK. Higher is better)}
    \label{fig:rmse-sched}
\end{figure}

\begin{figure}[t]
    \centering
    \includegraphics[width=0.55\textwidth]{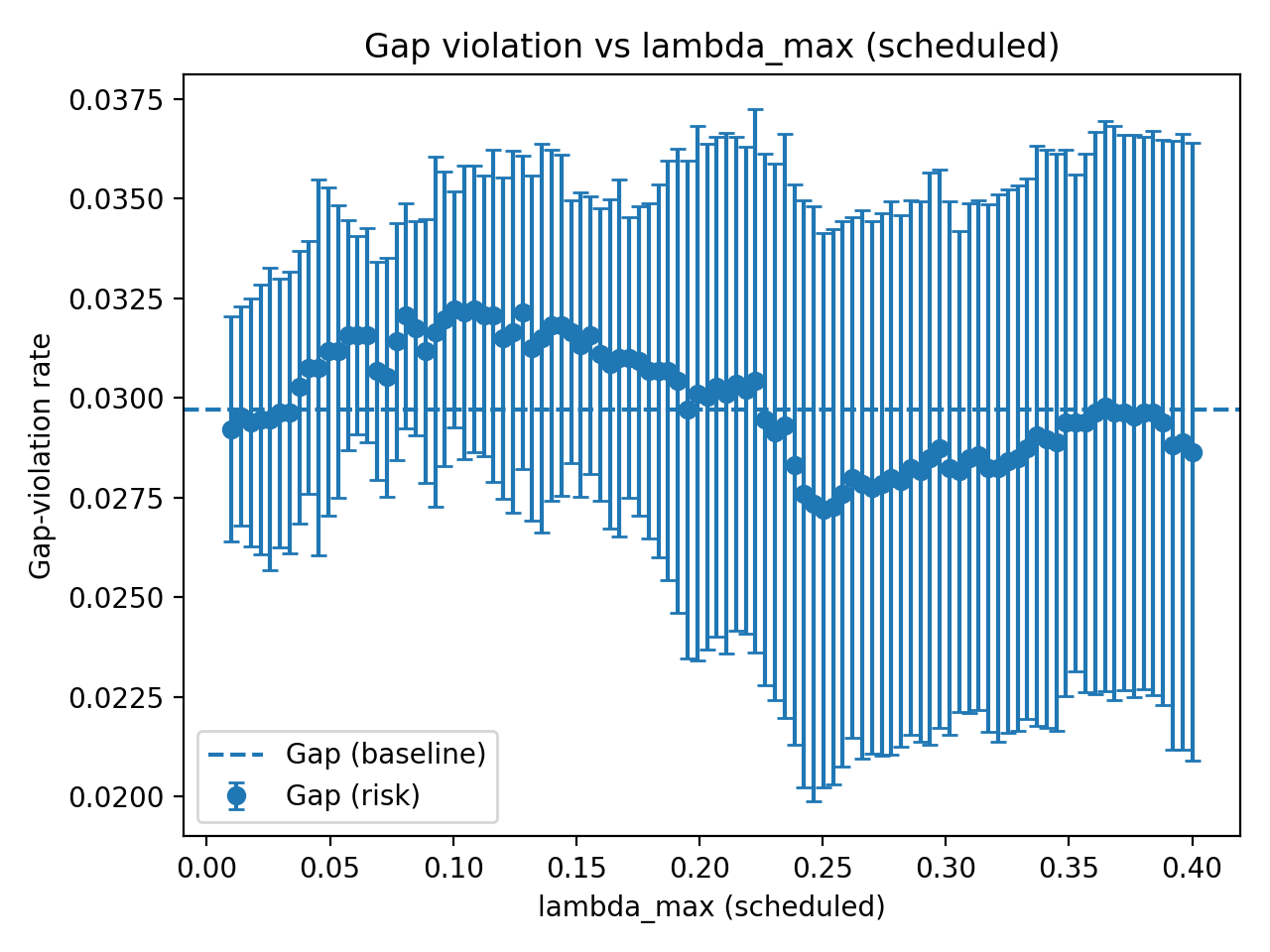}
    \caption{Gap violation rate vs.\ $\lambda_{\max}$ (FM baseline (dashed line) vs.\ scheduled FM-RISK. Lower is better).}
    \label{fig:gap-sched}
\end{figure}

\begin{figure}[t]
    \centering
    \includegraphics[width=0.55\textwidth]{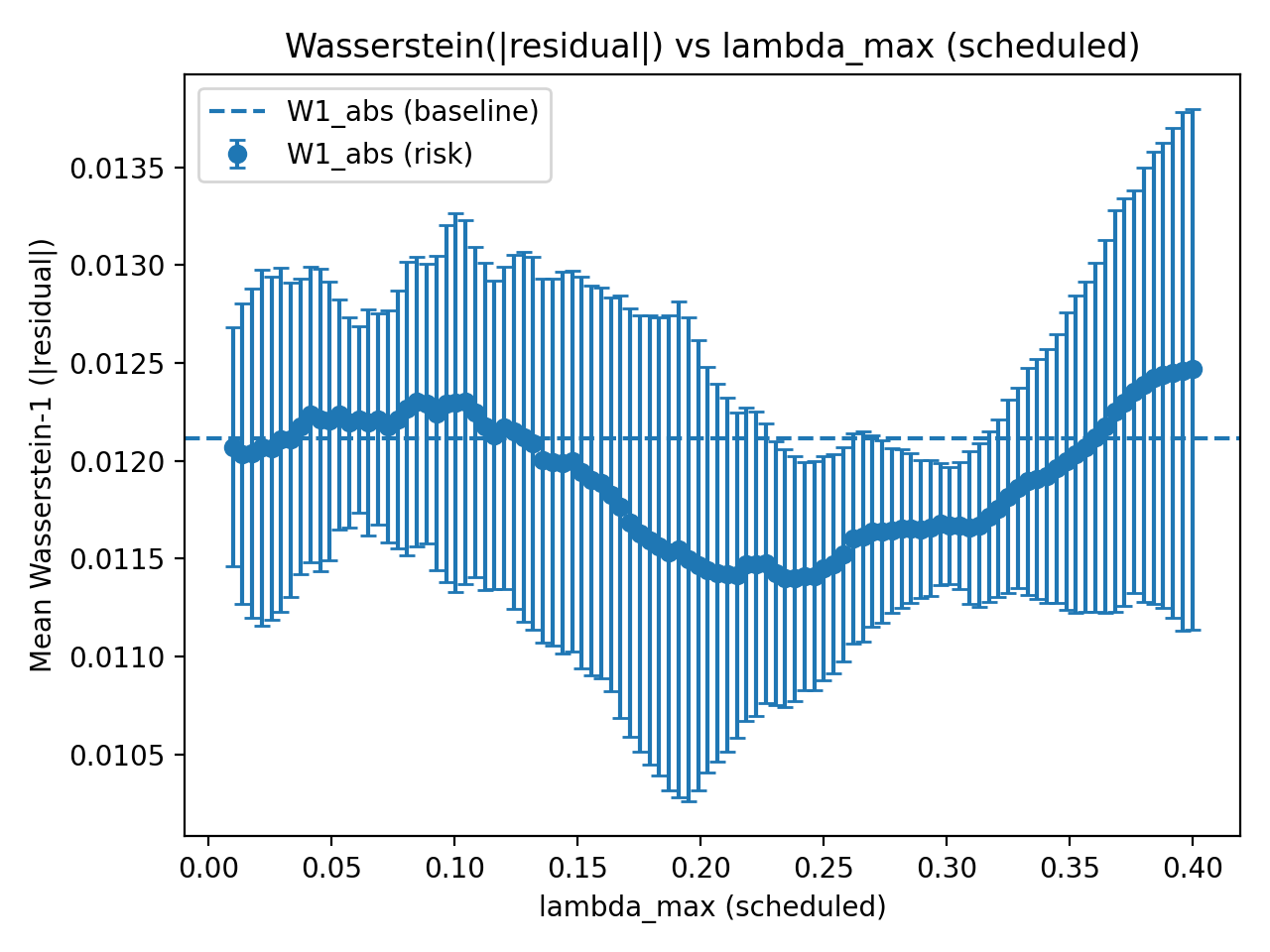}
    \caption{Mean Wasserstein--1 distance on absolute angular residuals
    $\mathrm{W}_1(|r|)$ vs.\ $\lambda_{\max}$ (FM baseline (dashed line) vs.\ scheduled FM-RISK. Lower is better).}
    \label{fig:wasserstein-sched}
\end{figure}

\paragraph{Results (scheduled risk sweep).}
Across a sweep of $\lambda_{\max}\!\in\![0.01,0.40]$, the entropic objective
consistently improves \emph{variance matching} and \emph{mode concentration}
while largely preserving mass away from inter-lobe gaps.
The plain FM baseline (corresponding to $\lambda_{\max}=0$) attains
\[
\mathrm{RMSE}_\sigma^{\text{base}} \approx 0.0149,\qquad
\mathrm{W}_1^{\text{base}}(|r|) \approx 0.0121,\qquad
\mathrm{Gap}^{\text{base}} \approx 0.0297.
\]
As shown in Figure~\ref{fig:rmse-sched-raw}, $\mathrm{RMSE}_\sigma$ decreases
monotonically as $\lambda_{\max}$ increases. Figure~\ref{fig:rmse-sched} summarizes the corresponding
relative improvement. For a representative mid-range
setting $\lambda_{\max}\approx 0.25$ we already obtain a reduction of about
$31\%$ in $\mathrm{RMSE}_\sigma$ relative to the baseline, while the
Wasserstein distance Figure~\ref{fig:wasserstein-sched} improves by roughly $6\%$ and the gap violation rate  Figure~\ref{fig:gap-sched}
drops by about $8.5\%$.

At the upper end of the sweep
($\lambda_{\max}\approx 0.40$) the model reaches
$\mathrm{RMSE}_\sigma^{\text{risk}} \approx 0.0079$, corresponding to a
relative improvement of roughly $47\%$ over FM.  In this more aggressive
regime the gap rate remains slightly better than baseline, whereas
$\mathrm{W}_1(|r|)$ drifts back to around the FM value and becomes
marginally worse (about $3\%$ larger), reflecting a trade-off in which the
risk term prioritizes shrinking the core angular spread.

Figure~\ref{fig:gap-sched} reports the gap violation rate as a function of
$\lambda_{\max}$.  For small coefficients, the behavior is close to FM, but for
moderate values the violation rate clearly drops below the baseline, indicating
better respect of the underlying angular structure; for very large coefficients
the gap rate slightly increases again but remains comparable to or better than
FM.

 In Figure~\ref{fig:wasserstein-sched} we plot the mean
Wasserstein distance $\mathrm{W}_1(|r|)$ between the residual distribution and
its reference.  The curve exhibits a broad U-shape: moderate risk coefficients
reduce $\mathrm{W}_1(|r|)$ relative to FM, while extremely large values tilt
the objective enough that the distance grows back toward the baseline.

\paragraph{Joint improvement regime.}

Across the sweep there is a broad regime of \emph{moderate} risk coefficients in
which all three key angular metrics improve simultaneously relative to the FM
baseline. Empirically, for $\lambda_{\max}$ in the range
$\lambda_{\max} \approx 0.2$–$0.3$, we observe concurrent reductions in
$\mathrm{RMSE}_\sigma$, calibrated gap-violation rate, and
$\mathrm{W}_1(|r|)$: the angular spread error typically decreases by
roughly $30$–$40\%$, the gap rate drops by about $5$–$9\%$, and the
Wasserstein distance on absolute residuals improves by around $5$–$6\%$
relative to baseline.
Outside this band the risk-sensitive objective still yields gains
in $\mathrm{RMSE}_\sigma$, but the improvements in gap rate and
$\mathrm{W}_1(|r|)$ become weaker or trade off against each other, indicating
that moderate risk levels are best aligned with jointly sharpening lobe
concentration, reducing gap mass, and matching the residual distribution.

On this simple, "smooth" two-ring benchmark, standard flow matching already achieves, as expected, fairly small angular and gap errors, so the absolute improvements from risk-sensitive training are modest in magnitude, even though the relative reductions in spread error and gap rate are substantial.

\section{Limitations and Future Work}

Our method introduces an entropic (log-exp) regularizer for flow matching with a temperature $\lambda>0$. While the first-order analysis clarifies that the intended regime is small $\lambda$, we do not yet provide a principled rule for choosing or adapting $\lambda$.

\paragraph{Hyperparameter and scheduling.}
We currently set $\lambda$ by a coarse sweep and found that scheduling it to ramp up later in training helps as early iterations are noise-dominated, so higher order (beyond the conditional mean) information is unreliable. Studying the scheduling of the $\lambda$ hyperparameter remains an interesting problem.

\paragraph{Scope of experiments.}
Experiments here are on controlled synthetic families. Broader validation on image/audio/text domains, ablations across network sizes and optimizers, and comparisons to hard-example mining and curriculum schedules are needed to assess robustness and scalability.

\section{Conclusion}

Tilted (log-exponential) risk on a loss is by now a standard tool in statistics
and machine learning, used to emphasize rare or high loss events while
preserving a clean optimization structure. In this work, we apply this
well established construction directly to flow matching. Classical flow
matching can be viewed as solving, at each $(t,x)$, the $L^2$ projection
problem so that the target velocity field is the conditional mean of a richer
microscopic velocity. Our entropic objective takes this characterization as
its starting point and replaces the plain mean square aggregation by a
Gibbs weighted one: we apply the tilted (log-exponential) risk to the FM loss
and show that, in the FM setting, the resulting global objective is a natural
upper-bound on a conditional entropic FM functional defined at each
space-time point. This yields a one parameter family of risk entropic
flow matching. At small values of this coefficient or  
temperature $\lambda$ the objective behaves like standard FM plus a 
correction, offering robustness and sharper control of higher order statistics;
at higher values it might act as an explicit innovation or exploration
mechanism, gradually reweighting tails and rare modes and thereby changing the
effective measure on the data manifold in a principled way.

\section {Appendix A: Synthetic 2D ring experiment}
\label{app:2d-ring}

\paragraph{Distributions.}
We consider a pair of 2D distributions supported on concentric circles with six
modes each.
The source distribution $p_0$ is defined as an equally weighted mixture of
$K = 6$ Gaussian components with means
\[
  m_k^{(0)} = r_{\mathrm{src}}
  \begin{bmatrix}
    \cos(2\pi k / K) \\
    \sin(2\pi k / K)
  \end{bmatrix},
  \qquad k \in \{0,\dots, K-1\},
\]
where $r_{\mathrm{src}} = 1/3$.
To sample from $p_0$, we first draw a component index $k$ uniformly from
$\{0,\dots,K-1\}$, then add independent angular and radial perturbations.
Concretely, if $\alpha_k = 2\pi k / K$ denotes the base angle, we sample
\[
  \tilde{\alpha} \sim \mathcal{N}(\alpha_k, \sigma_{\mathrm{ang}}^2),
  \qquad
  \tilde{r} \sim \mathcal{N}(r_{\mathrm{src}}, \sigma_{\mathrm{rad}}^2),
\]
with $\sigma_{\mathrm{ang}} = 0.12$ and $\sigma_{\mathrm{rad}} = 0.02$, and set
\[
  x_0 = \tilde{r}
  \begin{bmatrix}
    \cos \tilde{\alpha} \\
    \sin \tilde{\alpha}
  \end{bmatrix}.
\]
The target distribution $p_1$ is constructed analogously, but with radius
$r_{\mathrm{tgt}} = 1$, i.e.,
\[
  m_k^{(1)} = r_{\mathrm{tgt}}
  \begin{bmatrix}
    \cos(2\pi k / K) \\
    \sin(2\pi k / K)
  \end{bmatrix}.
\]
Angular and radial noise use the same standard deviations
$\sigma_{\mathrm{ang}} = 0.12$ and $\sigma_{\mathrm{rad}} = 0.02$, so the source and target differ only by overall scale.

\paragraph{Training tuples and baseline FM.}
At each optimization step we draw a mini batch of $B = 1000$ i.i.d.\ pairs
$(x_0, x_1)$, where $x_0 \sim p_0$ and $x_1 \sim p_1$ are sampled independently
as described above, together with times $t \sim \mathrm{Unif}[0, 1]$.
We form rectified paths
\[
  x_t = (1 - t)\,x_0 + t\,x_1,
  \qquad
  u = x_1 - x_0,
\]
and train a neural velocity field $v_\theta(x_t,t)$ using the standard
rectified flow matching loss
\[
  \mathcal{L}_{\mathrm{FM}}(\theta)
  = \mathbb{E}\,\bigl[\|v_\theta(x_t, t) - u\|_2^2\bigr].
\]
The baseline FM model is trained for $20{,}000$ iterations with batch size
$B=1000$, AdamW optimizer, learning rate $3 \times 10^{-4}$, and default weight
decay.

\paragraph{Risk-sensitive objective.}
For the risk-sensitive variant we keep the same data construction,
architecture, optimizer, and training schedule, changing only the loss.
Let
\[
  \ell_b(\theta) = \|v_\theta(x_t^{(b)}, t^{(b)}) - u^{(b)}\|_2^2
\]
denote the per-sample squared error for batch index $b \in \{1,\dots,B\}$.
For a risk coefficient $\lambda > 0$, the empirical log-exponential loss is
\[
  \hat{\mathcal{L}}_\lambda(\theta)
  = \frac{1}{\lambda}
    \log\left(
      \frac{1}{B}
      \sum_{b=1}^B
      \exp\bigl(\lambda\,\ell_b(\theta)\bigr)
    \right),
\]
which implements the entropic risk
$\frac{1}{\lambda}\log \mathbb{E}[\exp(\lambda \ell)]$ via a numerically stable
log-mean-exp over the mini-batch.
As $\lambda \to 0$ this reduces to the MSE loss.

We use a simple schedule $\lambda_t$ that ramps the risk coefficient from $0$
to a prescribed $\lambda_{\max}$ over the first $500$ training iterations and then keeps
it fixed:
\[
  \lambda_t =
  \begin{cases}
    0, & t < 0, \\[3pt]
    \frac{t}{500}\,\lambda_{\max}, & 0 \le t < 500, \\[3pt]
    \lambda_{\max}, & t \ge 500.
  \end{cases}
\]

For each value of $\lambda_{\max}$ in a sweep we retrain the model from scratch
with random uniform initialization and data seeds to enable paired comparisons
against the baseline FM model.

\paragraph{Per mode angular spread metric.}
To quantify how well the learned flow preserves the multi-modal structure of the
target distribution, we measure the angular concentration of mass around each
mode on the ring. Let $K$ denote the number of lobes (here $K=6$), and define the ideal mode
centers on the unit circle by
\[
  \mu_k \;=\; \frac{2\pi k}{K}, 
  \qquad k \in \{0,\dots,K-1\}.
\]
Given a 2D-point $x = (x_1, x_2)^\top$, we first compute its polar angle
\[
  \theta(x) \;=\; \operatorname{atan2}(x_2, x_1) \in [0,2\pi).
\]
We then assign $x$ to its nearest lobe and work in \emph{per-lobe coordinates}
by subtracting the closest mode center:
\[
  k^\star(x) \;=\; \arg\min_{k} \bigl|\mathrm{wrap}(\theta(x) - \mu_k)\bigr|,
  \qquad
  \delta\theta(x) \;=\; \mathrm{wrap}\bigl(\theta(x) - \mu_{k^\star(x)}\bigr),
\]
where $\mathrm{wrap}(\cdot)$ maps angles into $(-\pi,\pi]$ by adding or
subtracting integer multiples of $2\pi$.
Intuitively, $\delta\theta(x)$ is the angular residual of $x$ relative to the
center of its nearest mode, so that all $K$ lobes are folded onto a single
canonical lobe.

We apply this construction to both the ground-truth samples $x_1$ and the model
generated samples $\hat{x}_1$, obtaining residuals
$\delta\theta_{\text{true}}$ and $\delta\theta_{\text{model}}$, respectively.
For each $K$ group we then compute the \emph{circular} standard deviation of
these residuals,
\[
  \sigma_{\text{true}} \;=\; \operatorname{CircStd}\bigl(\delta\theta_{\text{true}}\bigr),
  \qquad
  \sigma_{\text{model}} \;=\; \operatorname{CircStd}\bigl(\delta\theta_{\text{model}}\bigr),
\]
using the resultant-length-based definition of circular standard deviation.
This measures the angular spread of mass around the nearest mode in a way that
is consistent on the circle.

\subsection{Metrics and Summary Statistics}

\paragraph{Gap violation rate (calibrated).}
Let $K$ be the number of angular lobes and let the lobe centers be
$\mu_j = 2\pi j/K$, $j=0,\dots,K-1$.
For a predicted angle $\theta \in [0,2\pi)$ define the wrapped distance to the
nearest lobe
\[
  d_K(\theta) \;=\; \min_{j}\ \big|\,\mathrm{wrap}(\theta-\mu_j)\,\big|,
  \quad
  \mathrm{wrap}(\phi)\in(-\pi,\pi].
\]
For each $K$ we first \emph{calibrate} a threshold $\tau_K$ from the
ground-truth residuals by choosing a high quantile (e.g.\ the $99$th
percentile) of $|r|$ for that $K$.
A model prediction with angle $\theta$ is a \emph{gap violation} if
$d_K(\theta) > \tau_K$; samples beyond this calibrated angular distance lie in
inter-lobe regions that are rarely visited by the true distribution.
Given a set of predictions $\{\theta_i\}_{i=1}^N$, the overall rate is
\[
  \mathrm{GapRate}
  \;=\;
  \frac{1}{N}\sum_{i=1}^N \mathbb{I}\!\left[d_K(\theta_i)>\tau_{K(i)}\right],
\]
where $K(i)$ denotes the lobe multiplicity associated with $\theta_i$ (here
always $K=6$).
Lower is better, as it indicates less mass in gap regions, and the use of
$\tau_K$ ensures that the notion of a “gap” is calibrated to the ground-truth
spread.

\paragraph{Wasserstein distances on angular residuals.}
For a given $K$, define nearest-lobe angular residuals for ground truth and
model by
\[
  r^{\text{true}}_i
  \;=\;
  \mathrm{wrap}\bigl(\theta^{\text{true}}_i-\mu_{j^\star(\theta^{\text{true}}_i)}\bigr),
  \qquad
  r^{\text{model}}_i
  \;=\;
  \mathrm{wrap}\bigl(\theta^{\text{model}}_i-\mu_{j^\star(\theta^{\text{model}}_i)}\bigr).
\]
We consider the empirical distributions of $|r|$ and of signed $r$ and compute
their one-dimensional Wasserstein--1 distances:
\[
  \mathrm{W}_1^{(K)}(|r|)
  \;=\;
  W_1\!\bigl(\{|r^{\text{true}}_i|\}_i,\{|r^{\text{model}}_i|\}_i\bigr),
\]
 \[
  \mathrm{W}_1^{(K)}(r)
  \;=\;
  W_1\!\bigl(\{r^{\text{true}}_i\}_i,\{r^{\text{model}}_i\}_i\bigr),
\]
where $W_1$ denotes the usual earth-mover distance on $\mathbb{R}$.
We then average these per-$K$ distances with weights proportional to the number
of samples in each bin:
\[
  \mathrm{W}_1(|r|)
  \;=\;
  \sum_K w_K\,\mathrm{W}_1^{(K)}(|r|),
  \qquad
  \mathrm{W}_1(r)
  \;=\;
  \sum_K w_K\,\mathrm{W}_1^{(K)}(r),
\]
with $\sum_K w_K = 1$.
Lower values indicate that the model reproduces the shape of the residual
distribution more closely, both in magnitude and in signed orientation.

\paragraph{Spread error \texorpdfstring{$\mathrm{RMSE}_\sigma$}{RMSE\_sigma}.}
Let $\sigma_{\text{true}}(K)$ be the circular standard deviation of the
nearest-lobe residuals for ground truth and $\sigma_{\text{model}}(K)$ for the
model.  With weights $w_K$ (again proportional to per-bin counts),
\[
  \mathrm{RMSE}_\sigma
  \;=\;
  \left(
    \sum_{K} w_K \big(\sigma_{\text{model}}(K)-\sigma_{\text{true}}(K)\big)^2
  \right)^{1/2}.
\]
This summarizes how well the model matches the angular spread of each lobe;
lower is better.

\paragraph{Radial error metrics.}
To ensure that improvements in angular statistics do not come at the expense of
radial transport, we also track radial mean-squared and mean-absolute errors.
Let $r^{\text{true}}_i = \|x^{(i)}_1\|_2$ and
$r^{\text{model}}_i = \|\hat{x}^{(i)}_1\|_2$ denote the radii of ground truth and model samples, respectively.
We define
\[
  \mathrm{RadialMSE}
  \;=\;
  \frac{1}{N}\sum_{i=1}^N \bigl(r^{\text{model}}_i - r^{\text{true}}_i\bigr)^2,
  \]
  \[
  \mathrm{RadialMAE}
  \;=\;
  \frac{1}{N}\sum_{i=1}^N \bigl|r^{\text{model}}_i - r^{\text{true}}_i\bigr|.
\]
These metrics are reported alongside the angular diagnostics but play a
supporting role: in our experiments, risk-sensitive training improves angular
behaviour while leaving radial errors essentially unchanged or slightly better.

\section*{Appendix B: Tilted log-exponential gradients}

In this appendix we give a more detailed derivation of the gradient of the conditional
log-exponential (entropic) FM objective and its first-order expansion in the
tilt parameter $\lambda$. Throughout, $\lambda > 0$ is a scalar parameter.
The case $\lambda=0$ recovers the usual FM/MSE objective; small positive
values of $\lambda$ define a deformation of that objective. But first, we need a lemma:

\begin{lemma}[First-order ratio expansion]
\label{lem:ratio-expansion}
Let
\[
N(\lambda) = N_0 + \lambda N_1 + O(\lambda^2), 
\qquad
D(\lambda) = D_0 + \lambda D_1 + O(\lambda^2),
\]
with $D_0 \neq 0$. Then
\[
\frac{N(\lambda)}{D(\lambda)}
=
\frac{N_0}{D_0}
+
\lambda\,\frac{N_1 D_0 - N_0 D_1}{D_0^2}
+
O(\lambda^2).
\]
\end{lemma}

\begin{proof}
Starting from
\[
D(\lambda) = D_0 + \lambda D_1 + O(\lambda^2),
\]
factor out $D_0$:
\[
D(\lambda)
=
D_0\Bigl(1 + \lambda \tfrac{D_1}{D_0} + O(\lambda^2)\Bigr).
\]
Set
\[
x(\lambda) := \lambda \tfrac{D_1}{D_0} + O(\lambda^2),
\]
so $x(\lambda) = O(\lambda)$ as $\lambda \to 0$. Then
\[
\frac{1}{D(\lambda)}
=
\frac{1}{D_0}\,\frac{1}{1 + x(\lambda)}.
\]
Using
\[
\frac{1}{1+x} = 1 - x + O(x^2) \quad \text{as } x \to 0,
\]
we obtain
\[
\frac{1}{1 + x(\lambda)}
=
1 - x(\lambda) + O(\lambda^2),
\]
and hence
\[
\frac{1}{D(\lambda)}
=
\frac{1}{D_0}
-
\lambda \frac{D_1}{D_0^2}
+
O(\lambda^2).
\]

With
\[
N(\lambda) = N_0 + \lambda N_1 + O(\lambda^2),
\]
we have
\begin{align*}
\frac{N(\lambda)}{D(\lambda)}
&=
\bigl(N_0 + \lambda N_1 + O(\lambda^2)\bigr)
\Bigl(
  \frac{1}{D_0}
  - \lambda \frac{D_1}{D_0^2}
  + O(\lambda^2)
\Bigr) \\
&=
N_0 \cdot \frac{1}{D_0}
+
\lambda\left(
  N_1 \cdot \frac{1}{D_0}
  - N_0 \cdot \frac{D_1}{D_0^2}
\right)
+
O(\lambda^2) \\
&=
\frac{N_0}{D_0}
+
\lambda\,\frac{N_1 D_0 - N_0 D_1}{D_0^2}
+
O(\lambda^2),
\end{align*}
which is the desired expansion.
\end{proof}

\begin{theorem}[First-order expansion of the tilted FM gradient]
\label{thm:tilted-FM-expansion}
Let $(X_t,t)$ denote the space--time pair arising from the rectified flow
matching construction, with law $\mathbb{P}_{X_t,t}$ (induced by the joint
law of the underlying data, e.g.\ $(X_0,X_1)$ and the sampling of $t$), and
assume this law is independent of the model parameters~$\theta$.

For each space--time point $(x,t)$, let $z$ be the conditional path data  variable attached to $(x,t)$, with conditional law $p(z \mid x,t)$, and
write
\[
  U_t(x,z) \in \mathbb{R}^d
\]
for the corresponding microscopic target velocity at $(x,t).$
Let
\[
  u_\theta^t(x) \in \mathbb{R}^d
\]
be a model prediction, differentiable in $\theta$.

Define the conditional covariance and third-order skew moment
\[
\Sigma_t(x)
:=
\mathbb{E}_{z\mid x,t}\big[\varepsilon_t(x,z)\,\varepsilon_t(x,z)^\top\big],
\qquad
S_t(x)
:=
\mathbb{E}_{z\mid x,t}\big[\varepsilon_t(x,z)\,\|\varepsilon_t(x,z)\|^2\big].
\]

For $\lambda > 0$, define the tilted local loss at $(x,t)$ by
\[
\ell_\lambda(x,t;\theta)
:=
\frac{1}{\lambda}\log
\mathbb{E}_{z\mid x,t}
\exp\!\big(\lambda \,\|u_\theta^t(x)-U_t(x,z)\|^2\big)
\quad (\lambda>0),
\]
with the convention
\[
\ell_0(x,t;\theta)
:=
\mathbb{E}_{z\mid x,t}
\big[\|u_\theta^t(x)-U_t(x,z)\|^2\big].
\]
Let the global tilted and FM/MSE objectives be
\[
L_\lambda(\theta)
:=
\mathbb{E}_{(X_t,t)}\big[\ell_\lambda(X_t,t;\theta)\big],
\qquad
\mathcal{L}_{\mathrm{MSE}}(\theta)
:=
\mathbb{E}_{(X_t,t)}\big[\ell_0(X_t,t;\theta)\big].
\]

Assume furthermore that
\begin{itemize}
\item for each $(x,t)$, the map $\theta \mapsto u_\theta^t(x)$ is $C^1$,
\item there exists an integrable function $G(x,t)$ such that
\[
\big\|\nabla_\theta u_\theta^t(x)\big\|
\;\le\; G(x,t)
\quad\text{for all } \theta \text{ in the parameter region of interest},
\]
and
\item the conditional moments
$\Sigma_t(x)$, $S_t(x)$ and the fourth moments of $\varepsilon_t(x,z)$ are
uniformly bounded on the support of $(X_t,t)$.
\end{itemize}

Then there exists $\lambda_0>0$ and a constant $C<\infty$ such that for all
$|\lambda|\le \lambda_0$,
\begin{align}
&\nabla_\theta L_\lambda(\theta)-\nabla_\theta \mathcal{L}_{\mathrm{MSE}}(\theta)
= \\&
2\,\mathbb{E}_{(X_t,t)}
\Big[
  \big(-\lambda\,S_t(X_t)+2\lambda\,\Sigma_t(X_t)\,m_t(X_t)\big)^\top
  \nabla_\theta u_\theta^t(X_t)
\Big]+R_\lambda(\theta),
\label{eq:tilted-FM-expansion}
\end{align}
where the remainder satisfies
\[
\big\|R_\lambda(\theta)\big\| \;\le\; C\,\lambda^2
\quad\text{for all } |\lambda|\le \lambda_0.
\]

\end{theorem}

\begin{proof}[Proof of Theorem]

Fix $(x,t)$ and write
\[
U := U_t(x,z), \qquad z \sim p(z\mid x,t),
\]
\[
u_\theta := u_\theta^t(x),
\]
so that $U$ is the random target and $u_\theta$ is the model prediction at
$(x,t)$.

For $\lambda\ge 0$ define
\begin{equation}
Z_\lambda(x,t;\theta)
:=
\mathbb{E}_{z\mid x,t}
\exp\!\big(\lambda \,\|u_\theta-U\|^2\big).
\label{eq:Zlambda-app}
\end{equation}
The local log-exponential loss is
\begin{equation}
\ell_\lambda(x,t;\theta)
:=
\frac{1}{\lambda}\log Z_\lambda(x,t;\theta)
\quad (\lambda>0),
\label{eq:elllambda-app}
\end{equation}
with the convention that $\ell_0$ is the usual squared loss. The global loss
is
\begin{equation}
L_\lambda(\theta)
:=
\mathbb{E}_{(X_t,t)}\big[\ell_\lambda(X_t,t;\theta)\big],
\label{eq:Llambda-app}
\end{equation}
where the sampling law of $(X_t,t)$ is fixed and does not depend on
$\theta$.

We first compute the gradient of $\ell_\lambda(x,t;\theta)$ at fixed $(x,t)$.
By the chain rule,
\begin{equation}
\nabla_\theta \ell_\lambda(x,t;\theta)
=
\frac{1}{\lambda}\,
\frac{\nabla_\theta Z_\lambda(x,t;\theta)}{Z_\lambda(x,t;\theta)}.
\label{eq:local-grad-1}
\end{equation}
Differentiating \eqref{eq:Zlambda-app} under the expectation and using
\[
\|u_\theta-U\|^2
= (u_\theta-U)^\top(u_\theta-U),
\qquad
\nabla_\theta \|u_\theta-U\|^2
=
2\,(u_\theta-U)^\top \nabla_\theta u_\theta,
\]
we obtain
\begin{align}
\nabla_\theta Z_\lambda(x,t;\theta)
&=
\mathbb{E}_{z\mid x,t}\Big[
  \exp\!\big(\lambda\|u_\theta-U\|^2\big)\,
  \lambda\,\nabla_\theta\|u_\theta-U\|^2
\Big] \notag\\
&=
2\lambda\,
\mathbb{E}_{z\mid x,t}\Big[
  \exp\!\big(\lambda\|u_\theta-U\|^2\big)\,
  (u_\theta-U)^\top
\Big]\nabla_\theta u_\theta.
\label{eq:dZ-app}
\end{align}
Substituting \eqref{eq:dZ-app} into \eqref{eq:local-grad-1} gives
\begin{equation}
\nabla_\theta \ell_\lambda(x,t;\theta)
=
2\,
\frac{
  \mathbb{E}_{z\mid x,t}\big[
    \exp(\lambda\|u_\theta-U\|^2)\,(u_\theta-U)^\top
  \big]
}{
  \mathbb{E}_{z\mid x,t}\big[
    \exp(\lambda\|u_\theta-U\|^2)
  \big]
}
\nabla_\theta u_\theta.
\label{eq:local-grad-2}
\end{equation}

Introduce the normalized weights
\begin{equation}
\beta_\lambda(z\mid x,t;\theta)
:=
\frac{
  \exp(\lambda\|u_\theta-U\|^2)
}{
  \mathbb{E}_{z'\mid x,t}
  \exp(\lambda\|u_\theta-U_t(x,z')\|^2)
},
\label{eq:beta-app}
\end{equation}
so that
\[
\mathbb{E}_{z\mid x,t}[\beta_\lambda(z\mid x,t;\theta)] = 1.
\]
Then the fraction in \eqref{eq:local-grad-2} can be written as
\[
\mathbb{E}_{z\mid x,t}\big[
  \beta_\lambda(z\mid x,t;\theta)\,(u_\theta-U)
\big]
=
u_\theta - \mathbb{E}_{z\mid x,t}\big[
  \beta_\lambda(z\mid x,t;\theta)\,U
\big].
\]
We define the tilted conditional mean
\begin{equation}
m_\lambda^t(x;\theta)
:=
\mathbb{E}_{z\mid x,t}\big[
  \beta_\lambda(z\mid x,t;\theta)\,U_t(x,z)
\big],
\label{eq:mtilt-app}
\end{equation}
so that
\[
\mathbb{E}_{z\mid x,t}\big[
  \beta_\lambda(z\mid x,t;\theta)\,(u_\theta-U)
\big]
=
u_\theta - m_\lambda^t(x;\theta).
\]
Equation \eqref{eq:local-grad-2} then becomes
\begin{equation}
\nabla_\theta \ell_\lambda(x,t;\theta)
=
2\,(u_\theta-m_\lambda^t(x;\theta))^\top
\nabla_\theta u_\theta.
\label{eq:local-grad-final}
\end{equation}

Taking the expectation in \eqref{eq:Llambda-app} and using that the
distribution of $(X_t,t)$ does not depend on $\theta$, we obtain the global
gradient identity
\begin{equation}
\nabla_\theta L_\lambda(\theta)
=
2\,\mathbb{E}_{(X_t,t)}\Big[
  \big(u_\theta^t(X_t)-m_\lambda^t(X_t;\theta)\big)^\top
  \nabla_\theta u_\theta^t(X_t)
\Big].
\label{eq:logexp-grad}
\end{equation}

For comparison, the FM/MSE objective has gradient
\begin{equation}
\nabla_\theta \mathcal{L}_{\mathrm{MSE}}(\theta)
=
2\,\mathbb{E}_{(X_t,t)}\Big[
  \big(u_\theta^t(X_t)-\mu_t(X_t)\big)^\top
  \nabla_\theta u_\theta^t(X_t)
\Big],
\label{eq:mse-grad3}
\end{equation}
where $\mu_t(x):=\mathbb{E}_{z\mid x,t}[U_t(x,z)]$ is the usual conditional
mean. Subtracting \eqref{eq:mse-grad3} from \eqref{eq:logexp-grad} and
rearranging gives the exact difference
\begin{equation}
\nabla_\theta L_\lambda(\theta)-\nabla_\theta \mathcal{L}_{\mathrm{MSE}}(\theta)
=
2\,\mathbb{E}_{(X_t,t)}\Big[
  \big(\mu_t(X_t)-m_\lambda^t(X_t;\theta)\big)^\top
  \nabla_\theta u_\theta^t(X_t)
\Big].
\label{eq:grad-diff-exact}
\end{equation}

\paragraph{Moment expansion of the tilted mean.}
We now expand $m_\lambda^t(x;\theta)$ in powers of $\lambda$ at fixed
$(x,t)$. Write
\[
U_t(x,z)=\mu_t(x)+\varepsilon,
\qquad \mathbb{E}[\varepsilon\mid x,t]=0,
\]
and define the conditional moments
\[
\Sigma_t(x)=\mathbb{E}[\varepsilon\,\varepsilon^\top\mid x,t],
\qquad
S_t(x)=\mathbb{E}[\varepsilon\,\|\varepsilon\|^2\mid x,t].
\]
Let the residual error be $m_t(x)=u_\theta^t(x)-\mu_t(x)$; then
\begin{equation}
\|u_\theta^t(x)-U_t(x,z)\|^2
=
\|m_t(x)-\varepsilon\|^2
=
\|m_t(x)\|^2 + \|\varepsilon\|^2 - 2\,m_t(x)^\top \varepsilon.
\label{eq:norm-expansion-app}
\end{equation}
Introduce
\begin{equation}
a := \|\varepsilon\|^2 - 2\,m_t(x)^\top \varepsilon.
\label{eq:a-app}
\end{equation}
The common factor $\exp(\lambda\|m_t(x)\|^2)$ cancels when forming the
normalized weights, so from \eqref{eq:mtilt-app} we may equivalently write
\begin{equation}
m_\lambda^t(x;\theta)
=
\frac{
  \mathbb{E}_{z\mid x,t}\big[(\mu_t(x)+\varepsilon)\,e^{\lambda a}\big]
}{
  \mathbb{E}_{z\mid x,t}\big[e^{\lambda a}\big]
}.
\label{eq:mtilt-fraction-app}
\end{equation}

For small $\lambda$ we expand
\begin{equation}
e^{\lambda a} = 1 + \lambda a + O(\lambda^2),
\label{eq:exp-expand-app}
\end{equation}
where $O(\lambda^2)$ denotes terms whose norm is bounded by a constant times
$\lambda^2$ for $\lambda$ in a fixed small interval; in particular such terms
vanish faster than $\lambda$ as $\lambda\downarrow 0$.

Using \eqref{eq:exp-expand-app}, the numerator
\[
N(\lambda)
:=
\mathbb{E}_{z\mid x,t}\big[(\mu_t(x)+\varepsilon)\,e^{\lambda a}\big]
\]
satisfies
\begin{align*}
N(\lambda)
&=
\mathbb{E}[\mu_t(x)+\varepsilon]
+
\lambda\,\mathbb{E}[(\mu_t(x)+\varepsilon)a]
+
O(\lambda^2) \\
&=
\mu_t(x)
+
\lambda\big(\mu_t(x)\,\mathbb{E}[a] + \mathbb{E}[\varepsilon a]\big)
+
O(\lambda^2),
\end{align*}
since $\mathbb{E}[\varepsilon\mid x,t]=0$. Similarly, the denominator
\[
D(\lambda)
:=
\mathbb{E}_{z\mid x,t}[e^{\lambda a}]
\]
has the expansion
\[
D(\lambda)
=
1 + \lambda\,\mathbb{E}[a] + O(\lambda^2).
\]

We can now use the elementary first-order ratio expansion
\begin{equation}
\frac{N_0+\lambda N_1+O(\lambda^2)}{D_0+\lambda D_1+O(\lambda^2)}
=
\frac{N_0}{D_0}
+
\lambda\,\frac{N_1 D_0 - N_0 D_1}{D_0^2}
+
O(\lambda^2),
\label{eq:ratio-app}
\end{equation}
which follows by expanding $1/(D_0+\lambda D_1)$ to first-order and
multiplying out. In our case,
\[
N_0=\mu_t(x),\quad
N_1=\mu_t(x)\,\mathbb{E}[a]+\mathbb{E}[\varepsilon a],
\quad
D_0=1,\quad
D_1=\mathbb{E}[a],
\]
so \eqref{eq:ratio-app} yields
\begin{align}
m_\lambda^t(x;\theta)
&=
\mu_t(x) + \lambda\big(N_1-\mu_t(x)D_1\big) + O(\lambda^2) \notag\\
&=
\mu_t(x) + \lambda\,\mathbb{E}[\varepsilon a] + O(\lambda^2).
\label{eq:mtilt-expand-1-app}
\end{align}
From \eqref{eq:a-app},
\[
\varepsilon a
=
\varepsilon\,\|\varepsilon\|^2
-
2\,\varepsilon\,\varepsilon^\top m_t(x),
\]
so
\begin{align*}
\mathbb{E}_{z\mid x,t}[\varepsilon a]
&=
\mathbb{E}[\varepsilon\,\|\varepsilon\|^2\mid x,t]
-
2\,\mathbb{E}[\varepsilon\,\varepsilon^\top\mid x,t]\,m_t(x) \\
&=
S_t(x)-2\,\Sigma_t(x)\,m_t(x).
\end{align*}
Substituting into \eqref{eq:mtilt-expand-1-app} gives the moment expansion
\begin{equation}
m_\lambda^t(x;\theta)
=
\mu_t(x)+\lambda\,S_t(x)-2\lambda\,\Sigma_t(x)\,m_t(x)+O(\lambda^2).
\label{eq:tilted-expansion1}
\end{equation}

\paragraph{First-order correction to the gradient.}
Finally we combine \eqref{eq:tilted-expansion1} with the exact gradient
difference \eqref{eq:grad-diff-exact}. At fixed $(x,t)$,
\[
\mu_t(x)-m_\lambda^t(x;\theta)
=
-\lambda\,S_t(x)+2\lambda\,\Sigma_t(x)\,m_t(x)+O(\lambda^2),
\]
and so
\begin{align}
&\nabla_\theta L_\lambda(\theta)-\nabla_\theta \mathcal{L}_{\mathrm{MSE}}(\theta)
=
2\,\mathbb{E}_{(X_t,t)}\Big[
  \big(\mu_t(X_t)-m_\lambda^t(X_t;\theta)\big)^\top
  \nabla_\theta u_\theta^t(X_t)
\Big] \notag\\
&=
2\,\mathbb{E}_{(X_t,t)}\Big[
  \big(-\lambda\,S_t(X_t)+2\lambda\,\Sigma_t(X_t)\,m_t(X_t)\big)^\top
  \nabla_\theta u_\theta^t(X_t)
\Big]
+O(\lambda^2).
\label{eq:delta-final1}
\end{align}

Thus, for small $\lambda$, the entropic FM gradient is the FM/MSE gradient
plus a first-order correction consisting of:
\begin{itemize}\setlength{\itemsep}{2pt}
\item a covariance-preconditioning term $2\lambda\,\Sigma_t(x)\,m_t(x)$, and
\item a skew/tail term $-\lambda\,S_t(x)$ that depends only on the data
  moments and does not involve the current residual.
\end{itemize}
The $O(\lambda^2)$ remainder term collects all higher-order contributions and
vanishes faster than $\lambda$ as $\lambda$ approaches~$0$.

\end{proof}

\paragraph{Gradient transfer to the marginal FM objective.}
Define the marginal FM loss
\begin{align*}
L_{\mathrm{FM}}(\theta)
&=
\mathbb{E}
\bigl[
  \|u_\theta^t(X_t) - u_t(X_t)\|^2
\bigr],
\end{align*}

with conditional mean
\begin{align*}
\mu_t(x)
&=
\mathbb{E}_{z \mid x,t}
\bigl[
  U_t(x,z)
\bigr].
\end{align*}

\begin{corollary}
\emph{Immediate corollary (by substitution).}
As shown in \cite{lipman2022flowmatching},
\begin{align*}
\nabla_\theta L_{\mathrm{FM}}(\theta)
&=
\nabla_\theta L_{\mathrm{MSE}}(\theta),
\end{align*}
which togther with \ref{eq:delta-final1} implies the same first–order correction applies to the marginal FM gradient:
\[
\nabla_\theta L_{\mathrm{FM}}(\theta)
\;\longrightarrow\;
\nabla_\theta L_{\mathrm{FM}}(\theta)
+
2\,\mathbb{E}
\Big[
  \bigl(
    -\lambda\,S_t(X_t)
    + 2\lambda\,\Sigma_t(X_t)\,m_t(X_t)
  \bigr)^\top
  \nabla_\theta u_\theta^t(X_t)
\Big].
\]\hfill$\square$
\end{corollary}

Remarkably, definitions of 
$S$ and $m_t$ imply that if the local velocity fields are symmetric and the baseline model perfectly  estimates the 
FM velocity field, the first order correction terms disappear; i.e, The \textit{original} FM optimum is a fixed point of our risk-sensitive / entropic deformation under those assumptions.

\end{document}